\documentclass[a4paper,twoside]{article}

\usepackage{epsfig}
\usepackage{subcaption}
\usepackage{calc}
\usepackage{amssymb}
\usepackage{amstext}
\usepackage{amsmath}
\usepackage{amsthm}
\usepackage{multicol}
\usepackage{pslatex}
\usepackage{apalike}
\usepackage{algorithm2e}
\usepackage[bottom]{footmisc}

\usepackage{algorithm}
\usepackage{soul}
\usepackage{makecell}
\usepackage{booktabs}
\usepackage{subcaption}
\usepackage{multicol}
\usepackage{multirow}
\usepackage{tcolorbox}
\usepackage{enumitem}
\usepackage{url}

\usepackage{SCITEPRESS}     

\begin{document}

\title{Rethinking Post-Training Quantization: Introducing a Statistical Pre-Calibration Approach}


\author{\authorname{Alireza Ghaffari\sup{1}, Sharareh Younesian\sup{2}, Boxing Chen\sup{2}, Vahid Partovi Nia\sup{2}, and Masoud Asgharian\sup{1}}
\affiliation{\sup{1}Department of Mathematics and Statistics, McGill University, Montreal Canada.}
\affiliation{\sup{2}Huawei Noah's Ark Lab, Montreal, Canada.}
\email{alireza.ghaffari@mcgill.ca}
}


\abstract{As Large Language Models (LLMs) become increasingly computationally complex, developing efficient deployment strategies, such as quantization, becomes crucial. State-of-the-art Post-training Quantization (PTQ) techniques often rely on calibration processes to maintain the accuracy of these models. However, while these calibration techniques can enhance performance in certain domains, they may not be as effective in others. This paper aims to draw attention to robust statistical approaches that can mitigate such issues. We propose a\textit{ weight-adaptive} PTQ method that can be considered a precursor to calibration-based PTQ methods, guiding the quantization process to preserve the distribution of weights by minimizing the Kullback-Leibler divergence between the quantized weights and the originally trained weights. This minimization ensures that the quantized model retains the Shannon information content of the original model to a great extent, guaranteeing robust and efficient deployment across many tasks. As such, our proposed approach can perform on par with most common calibration-based PTQ methods, establishing a new pre-calibration step for further adjusting the quantized weights with calibration. We show that our pre-calibration results achieve the same accuracy as some existing calibration-based PTQ methods on various LLMs.}

\onecolumn \maketitle \normalsize \setcounter{footnote}{0} \vfill

\section{\uppercase{Introduction}}
\label{sec:introduction}

Large Language Models (LLMs) have rapidly evolved, demonstrating unprecedented capabilities in natural language processing tasks. However, the immense computational resources required for their deployment pose significant challenges, particularly in resource-constrained environments. As these models become more complex, the need for efficient deployment strategies becomes increasingly critical. Quantization, a technique that reduces the precision of the model, has emerged as a promising solution to this problem by significantly reducing the computational and memory demands of LLMs while striving to maintain their performance.

Post-training quantization (PTQ) is a widely adopted approach for implementing quantization after a model has been fully trained. Traditionally, PTQ methods rely heavily on calibration processes to fine-tune the quantized model, ensuring that it retains a high degree of accuracy. These calibration techniques have proven effective in various domains, particularly when the target deployment environment closely resembles the conditions under which the model was calibrated. However, their efficacy may diminish in scenarios where the deployment environment diverges from the calibration conditions, leading to suboptimal performance.

For instance, Table \ref{tab:code} shows this limitation when quantizing a Code-Llama model using mainstream PTQ methods such as SpQR \cite{dettmers2024spqr}. To showcase the robustness issue of calibration-based PTQ method, we evaluated the coding performance of quantized Code-Llama model \cite{roziere2023code} on HumanEval \cite{chen2021codex} and MBPP \cite{austin2021program} datasets. HumanEval includes 164 human handwritten programming problems with a function signature, docstring, body, and several unit tests, and MBPP consists of around 1,000 crowd-sourced Python programming problems. 
Table \ref{tab:code} shows that a robust pre-calibration method outperforms SpQR\cite{dettmers2024spqr}, demonstrating that if calibration data does not have the same nature as the task, using calibration data decreases the performance.

\begin{table}[!t]
\caption{Comparison of \textit{weight-adaptive }pre-calibration results for Code-Llama models on HumanEval \cite{chen2021codex} and MBPP \cite{austin2021program}. }
    \centering
    \scalebox{0.47}{\begin{tabular}{ccccccc}
    \toprule
    Model     & Method & Avg Bits & \multicolumn{2}{c}{Human Eval} & \multicolumn{2}{c}{MBPP} \\
    \cmidrule(r){4-5} \cmidrule(r){6-7}
     & & & pass@1 & pass@10 & pass@1 & pass@10  \\
    \midrule
                & FP16  & 16.00 & 29.63 & 59.84 & 25.87 & 63.52  \\
                & RTN (g128)  & 4.25 & 30.13 & 57.97 & 28.26 & 62.42 \\
     Code-Llama-7B   & SpQR$^*$ & 4.63 & 29.94 & 57.40 & 27.59 & 61.78  \\
                & Pre-calibration (g128, $\alpha$=5\%) & 4.60 & \textbf{30.34} & \textbf{58.60} & 28.03 & \textbf{62.55} \\
    \midrule
               & FP16 &  16.00 & 34.79 & 66.50 & 30.17 & 67.51  \\
                & RTN (g128) & 4.25 & 33.70 & 65.88 & 29.63 & 66.00   \\
     Code-Llama-13B   & SpQR$^*$ & 4.63 & 34.19 & 65.69 & 29.74 & 66.20   \\
                & Pre-calibration (g128, $\alpha$=6\%) & 4.67 & \textbf{34.79} & \textbf{66.02} & \textbf{31.36} & \textbf{66.82} \\
    \bottomrule
  \end{tabular}
  }
  \label{tab:code}
  \end{table}

Given these limitations, there is a growing interest in exploring mathematical approaches to enhance the robustness of PTQ methods. In particular, statistical methods that guide the quantization process itself —prior to any calibration— offer a promising avenue for improvement. By focusing on preserving the underlying distribution of model weights, these approaches can potentially ensure a more consistent performance across diverse deployment scenarios.

In this paper, we introduce a novel \textit{ weight-adaptive pre-calibration} quantization method that functions as a precursor to traditional calibration-based techniques. Our method is grounded in a statistical framework that minimizes the Kullback-Leibler divergence between the original weights and quantized weights, thereby preserving the Shannon information content of the model. This pre-calibration step ensures that the quantized model remains robust across various tasks, even before any further calibration is applied.

Note that our approach not only preserves the accuracy of the quantized model but also sets a new initial point for subsequent calibration processes. Through extensive experiments on various LLMs, we show that our pre-calibration method achieves performance on par with existing calibration-based PTQ techniques, offering a more reliable and efficient deployment strategy for LLMs in diverse environments.

To summarize, we make the following contributions
\begin{itemize}
    \item  We introduce a weight-adaptive pre-calibration method that as a precursor to traditional calibration-based methods guides the quantization process to better preserve model information. To the best of our knowledge, this is the first time a statistical pre-calibration method has been proposed to improve the quantization process.
    \item Our proposed pre-calibration method \textit{classifies} weights and does not adjust them as opposed to traditional PTQ methods.
    The proposed method then uses pseudo activations (i.e. identity matrix) to identify and isolate important weights simplifying the algorithm to soft-thresholding which makes the pre-calibration computationally efficient.
    \item The proposed pre-calibration approach ensures that the quantized model performs consistently across a variety of deployment environments, addressing the limitations of calibration-based methods in domain-specific scenarios.
    \item Our work introduces a new pre-calibration step that can be integrated with existing PTQ calibration methods, offering a new initial point for the calibration optimization procedure, enhancing the overall effectiveness of the PTQ proces.
    \item We provide a theoretical foundation for our proposed pre-calibration method using information theory and techniques from statistical machine learning.
\end{itemize}

The rest of the paper is organized as follows. In Section \ref{sec:problem} we provide a detailed problem statement and clarify our proposed weight-adaptive pre-calibration. Section \ref{sec:related-works} reviews recent works in the field of PTQ and specifies the differences to our proposed weight-adaptive pre-calibration method. Section \ref{sec:method} discusses the proposed pre-calibration algorithm in detail. Section \ref{sec:theory} delves deeper into the theoretical analysis of the algorithm and shows how pre-calibration can control information loss in quantization. 
Finally, experimental results supporting our proposed methodology and theoretical findings are presented in Section \ref{sec:experiment}.

\section{Problem Statement}\label{sec:problem}
Recently proposed PTQ methods such as \cite{optq,quip} often use $\arg\min_{\hat{\textbf{W}}} \| \textbf{W}\textbf{X} - \hat{\textbf{W}}\textbf{X} \|_2^2$ to adjust the quantized model weights $\hat{\textbf{W}}$ with respect to original weights $\textbf{W}$ , ensuring that the reduction in precision does not significantly degrade performance.
In contrast, we propose a fundamentally different approach to PTQ notable as \textit{pre-calibration}, which re-frames the quantization process as a classification problem on the model's weights. This approach does not rely on any calibration data, setting it apart from the conventional PTQ methods. Instead of using calibration for post-hoc adjustments, our method classifies the model’s weights into quantization bins in a manner that inherently preserves the underlying distribution of the weights.

\subsection{Weight Adaptive Penalization}
Let us consider the following optimization problem

\begin{equation}
    \arg\min_{\hat{\textbf{W}}} \| \textbf{W}\textbf{X} - \hat{\textbf{W}}\textbf{X} \|_2^2 + \lambda \mathbb{D}_{\text{KL}}( f_{{\textbf{W}}}\| f_{\hat{\textbf{W}}}   ),
    \label{eq:obj_quant_KL}
\end{equation}
where ${\textbf{W}}$ denotes original weights with $f_{{\textbf{W}}}$ distribution  and $\hat{\textbf{W}}$ denotes quantized weights with   $f_{\hat{\textbf{W}}}$ distribution. 

Note that problem \eqref{eq:obj_quant_KL} is {\it only} used for classification, {\it not} shrinkage, of weights and the penalty term, $\lambda \mathbb{D}_{\text{KL}}( f_{{\textbf{W}}}\| f_{\hat{\textbf{W}}})$ , is used to guide the classification in a way that the distribution of quantized weights closely follows that of the original weights.

By viewing pre-calibration as a classification problem, we can ensure that the quantization process itself is robust, reducing the need for extensive calibration afterward. This method fundamentally shifts the focus from calibration after quantization to optimizing the quantization process from the outset, thereby enhancing the robustness and generalizability of the quantized model across various tasks.

\subsection{Weight Classification, Penalization,  and Saliency Detection}


Unlike traditional calibration-based methods that adjust quantized weights by solving 
$\arg\min_{\hat{\textbf{W}}} \| \textbf{W}\textbf{X} - \hat{\textbf{W}}\textbf{X} \|_2^2$
 , our proposed pre-calibration method does not modify the quantized weight tensor. Instead, the optimization problem \eqref{eq:obj_quant_KL} and its penalty term are employed solely to classify weights into two categories: salient weights and non-salient weights.

It is important to clarify that in this context, salient weights are not simply large values. Rather, our optimization framework defines salient weights as those that cause the distribution of quantized weights to deviate significantly from the original distribution. The penalty term 
$\lambda \mathbb{D}_{\text{KL}}( f_{{\textbf{W}}}\| f_{\hat{\textbf{W}}})$ is used specifically to ensure that the classification of weights is conducted in a way that it preserves the overall weight distribution after quantization.

\subsection{Pre-calibration and Pseudo Activations}
An inherent challenge that emerges from the optimization problem \eqref{eq:obj_quant_KL} is that activations $\textbf{X}$ are inherently tied to the input of a layer, implying a need for calibration. To address this issue, we remove the necessity for calibration by utilizing pseudo activations. For example, when $\textbf{XX}^{\top} = b\textbf{I}$, we can leverage specific mathematical properties to simplify the KL-divergence using a straightforward soft-thresholding approach.

\section{Related Works}\label{sec:related-works}

In the field of low-precision deep learning, three existing notable categories are (i) low-precision or quantized training, (ii) quantization-aware training (QAT), and (iii) post-training quantization (PTQ). While our proposed method can be applied to both low-precision training (e.g. \cite{banner2018scalable,zhang2020fixed,zhu2020towards,zhao2021distribution,ghaffari2022integer}) and QAT (e.g. \cite{zhu2023survey,dettmers2024qlora,liu2023llm} ), our primary focus is PTQ of LLMs which is found to be more challenging in the literature. As such, we confine our attention to PTQ of LLMs in this section.

Historically, PTQ methods were common for computer vision models with small number of parameters, some notable methods are AdaRound \cite{adaround}, OBQ \cite{obc}, AdaQuant \cite{adaquant}, and BRECQ \cite{brecq}. However, these methods were found to be either compute-intensive or inaccurate for large language models.

LLM.int8() \cite{llm-int8} and ZeroQuant \cite{zeroquant} are among the first PTQ techniques that were designed for LLMs. LLM.int8() separates the outlier activations and keeps them in floating-point number format while quantizing weights and non-outlier activations to 8-bit integers. LLM.int8() separates the outlier activations based on their magnitude. On the other hand, ZeroQuant uses a fine-grained hardware-friendly quantization scheme as well as layer-by-layer knowledge distillation for quantizing both weight and activations. However, both LLM.int8() and ZeroQuant are not efficient for quantizing LLMs to extreme low-precision number formats such as 3-bit integers.

OPTQ \cite{optq} is a PTQ algorithm for LLMs that can quantize weights to 3- or 4-bit integers. OPTQ adapted a calibration algorithm inspired by \cite{obs} that minimizes the $\ell_2$ loss of the quantized layer output with the original output. SpQR \cite{dettmers2024spqr} uses OPTQ algorithm while separating the salient weights and keeping them in FP16 format and further uses double quantization to reduce the memory. Both SpQR and OPTQ algorithms require calibration data for quantization.

SmoothQuant \cite{smoothquant} performs 8-bit integer quantization of weights and activation by offline migration of the quantization difficulty from activations to weights. Likewise, AWQ \cite{awq}, quantized weights by applying per-channel scales that protect the salient weights by observing the activation. SmoothQuant and AWQ algorithms also require calibration data to perform quantization. 

QuarRot \cite{ashkboos2024quarot}, AQLM \cite{aqlm} and QServe \cite{lin2024qserve} are among the most recent PTQ approaches. Quarot tackles the outlier problem by rotating LLMs in a way that removes outliers from the hidden state without changing the output. AQLM generalizes the classic Additive Quantization (AQ) approach for LLMs. QServe introduces a  quantization algorithm with 4-bit weight, 8-bit activation, and 4-bit KV cache. Moreover, QServe introduces SmoothAttention to effectively mitigate the accuracy degradation incurred by 4-bit KV quantization.

The key feature of our proposed pre-calibration algorithm lies in its ability to classify weights to improve quantization accuracy without performing any calibration. Furthermore, our proposed method uniquely classifies and isolates outlier weights solely through analysis of the model weight tensors ensuring more robust quantization that does not depend on the calibration dataset. While our proposed methodology is a precursor to PTQ algorithms, it can also be combined with calibration based method to improve the accuracy.

\section{Methodology}\label{sec:method}

The core intuition behind our approach is rooted in the goal of matching the distribution of quantized weights to that of the original weights as explained by optimization problem \eqref{eq:obj_quant_KL}. A straightforward way to achieve this is by ensuring that each quantized weight is as close as possible to its corresponding original weight i.e. ${\hat w_i}= {w_i}$ or $\frac{\hat w_i}{w_i}-1 =0 \text{~~s.t.~~~} w_i \neq 0$. This proximity naturally preserves the overall distribution, minimizing the divergence between the original and quantized weight distributions.

However, directly matching each quantized weight to its original counterpart may not always be possible, especially when the quantization process introduces significant changes in the weight values. To achieve parsimony, we may decide, according to the "importance" of each weight, how close a quantized weight should be matched with its original counterpart. Given this basic intuition, we are led to the classification of weights. 
The question then arises as to how such classification should be done. To this end, we consider penalization methods for classifications where the penalty on each quantized weight is gauged and guided by its original weight, the available gold standard. One penalty that serves such purpose well is called Adaptive LASSO \cite{zou2006adaptiveLASSO} in statistical machine learning literature. Adpative Lasso is the penalty of choice when a gold standard exists. 
\begin{equation}
    \arg\min_{\hat{\textbf{W}}} \| \textbf{W}\textbf{X} - \hat{\textbf{W}}\textbf{X} \|_2^2 + \lambda \sum_{i} \left |\frac{\hat w_i}{w_i} \right|,
    \label{eq:adaptive-lasso}
\end{equation}
The mathematical proof of how problem \eqref{eq:adaptive-lasso} is a proxy solution to problem \eqref{eq:obj_quant_KL} is presented in Section \ref{sec:theory}. We emphasize that the penalization method is {\it only} used for classification of weights into salient and non-salient, in statistics language active and inactive, weights, {\it not} for shrinkage. 


\subsection{Pseudo Activations}
A key challenge arising from the optimization problem \eqref{eq:adaptive-lasso} is that activations $\textbf{X}$ are intrinsically linked to the input of a layer, suggesting a requirement for calibration. To overcome this, we eliminate the need for calibration by employing pseudo activations. 
Let us assume $\textbf{X}$ is an orthogonal matrix i.e. $\textbf{X}\textbf{X}^\top=b\textbf{I}$, where $b$ is a constant and $\textbf{I}$ is the identity matrix. By expanding \eqref{eq:adaptive-lasso} we have
\begin{align}
      &\mathcal{L} = \big (\textbf{W}\textbf{X} - \hat{\textbf{W}}\textbf{X} \big ) \big (\textbf{W}\textbf{X} - \hat{\textbf{W}}\textbf{X} \big )^\top + \lambda \sum_{i} \left |\frac{\hat w_i}{w_i} \right|\\\nonumber
      &= b\|{\textbf{W}}\|_2^2 - 2b\hat{\textbf{W}} \textbf{W}^\top + b\|\hat{\textbf{W}}\|_2^2 + \lambda \sum_{i} \left|\frac{\hat w_i}{w_i}\right|,
      \label{eq:soft-th-dev1}
\end{align}
and since $\textbf{W}$, weights of the original model are constant, the Adaptive LASSO loss becomes
 \begin{align}
      & \mathcal{L} = -2b\hat{\textbf{W}} \textbf{W}^\top + b\|\hat{\textbf{W}}\|_2^2 + \lambda \sum_{i} \left|\frac{\hat w_i}{w_i}\right| \\\nonumber
      &= \sum_{i} \big ( -2b w_i \hat{w}_i + b\hat{w}_i^2 + \lambda \left |\frac{\hat{w}_i}{w_i} \right| \big ).
      \label{eq:soft-th-dev3}
\end{align}
By taking the derivative with respect to $\hat{w}_i$, and setting it equal to zero, it is easy to see
\begin{equation}
      \hat{w}_i = \text{sign}({w}_i) \text{{ReLU}}(|{w}_i|-\frac{\lambda'}{|{w}_i|}),
      \label{eq:soft-th}
\end{equation}
where $\lambda' = {\lambda}/{2b}$ and $\text{{ReLU}}()$ denotes the positive part i.e. $\text{{ReLU}}(x) = \max (x,0)$. Equation \eqref{eq:soft-th} shows that Adaptive LASSO is a simple soft-thresholding method that is very efficient to be implemented in the currently available commodity hardware.

\subsection{Proposed Pre-Calibration Algorithm}
To match the distributions of original and quantized weights using KL divergence, we employ adaptive lasso penalty term, and the combination of Adaptive LASSO with pseudo activations naturally leads to a soft-thresholding approach as shown in \eqref{eq:soft-th}.  Through using soft-thresholding for classifications of weights, we achieve a quantized model that not only retains the key characteristics of the original but also ensures robust performance across diverse tasks. Algorithm 1 shows this classification procedure. Note that in Algorithm \ref{alg:al_alg},  none of the weights are shrunk to zero and equation \eqref{eq:soft-th} is only used to classify weights to salient and non-salient weights. Here, salient weights are defined  as those weights that cause the distribution of quantized weights to deviate significantly from the original distribution.

\begin{algorithm}[H]{
\textbf{Input:} Layer weight tensor $\textbf{W}$, Outlier percentage $\alpha$\\
\textbf{Step 1.} Start from  a large $\lambda' >> 0$ in \eqref{eq:soft-th} then reduce it  until $\alpha$ percent of weights are selected as outlier. (Class 1) \\
\textbf{Step 2.} Classify all other weights as common weights. (Class 2)\\
\textbf{Step 3.} Quantize Class 1 weights and Class 2 weights using minmax quantization\\
\caption{Pre-Calibration algorithm}
\label{alg:al_alg}
}\end{algorithm}


\section{Theoretical Considerations}\label{sec:theory}

In this section, we establish the theoretical foundation that underpins our approach, specifically demonstrating that the Adaptive LASSO serves as a proxy solution to minimizing the KL divergence between the original and quantized weight distributions. By rigorously analyzing the relationship between adaptive lasso regularization and KL divergence, we show that the adaptive lasso effectively guides the quantization process toward preserving the original model’s weight distribution. 

Suppose $f_{\textbf{W}}$ is twice continuously differentiable. Let $f'_{\textbf{W}}$ and $f''_{\textbf{W}}$ denote the first and second derivatives of $f_{\textbf{W}}$. Suppose the mean $\mu_{\delta}$ and variance $\sigma^2_{\delta}$ of the quantization error $\delta$ are small. Then

\noindent\textbf{Claim 1:} 

\small{
$\mathbb{D}_{\text{KL}}( f_{{\textbf{W}}}\| f_{\hat{\textbf{W}}}) \approx \mu_{\delta} \sum_{i} f'_{{\textbf{W}}}(w_i) + \mu_{\delta} \sum_{i} f''_{{\textbf{W}}}( w_i) (\hat w_i - w_i)$
}

\noindent\textbf{Claim 2:}  

$\left |\mu_{\delta} \sum_{i} f''_{{\textbf{W}}}( w_i) (\hat w_i - w_i) \right| \leq C \left ( \sum_{i} \left |\frac{\hat w_i}{w_i} \right| + 1 \right )$ where $C$ is a constant.

\textbf{\textit{Proof:}} Assuming a quantization error $\delta_{i}$, each original weight relates to the quantized weight such that $\hat{w}_i = {w}_i + \delta_{i}$. Let us also assume errors $\delta$ are independent of the weights values. Therefore, quantized weight distribution is a convolution of original weights distribution and quantization error distribution such that
\begin{align}
    f_{\hat{\textbf{W}}}(\hat w) &= (f_{{\textbf{W}}} \ast f_{\delta})(\hat w) = 
    \int_{-\infty}^{\infty}f_{{\textbf{W}}}(\hat w-x)f_{\delta}(x)dx\\\nonumber
    &=f_{{\textbf{W}}}(\hat w) + \int_{-\infty}^{\infty}(f_{{\textbf{W}}}(\hat w-x)-f_{{\textbf{W}}}(\hat w))f_{\delta}(x)dx.
    \label{eq:conv_dist}
\end{align}
Using the mean value theorem for $f_{{\textbf{W}}}(\hat w-x)-f_{{\textbf{W}}}(\hat w)$, we have
\begin{align}
    f_{\hat{\textbf{W}}}(\hat w) \nonumber &=f_{{\textbf{W}}}(\hat w) + \int_{-\infty}^{\infty}(-x)f'_{{\textbf{W}}}(\xi_{\hat w}(x))f_{\delta}(x)dx\\
    & \stackrel{\sigma^2_{\delta} \text{ is small}}{\approx} f_{{\textbf{W}}}(\hat w) - \int_{-\infty}^{\infty}xf'_{{\textbf{W}}}(\hat w)f_{\delta}(x)dx, \\\nonumber
\end{align}
and thus,
\begin{align}
    &\frac{f_{\hat{\textbf{W}}}(\hat w)}{f_{{\textbf{W}}}(\hat w)} \approx 1 - \int_{-\infty}^{\infty}\frac{xf'_{{\textbf{W}}}(\hat w)}{f_{{\textbf{W}}}(\hat w)}f_{\delta}(x)dx\\\nonumber
    &= 1 - \frac{f'_{{\textbf{W}}}(\hat w)}{f_{{\textbf{W}}}(\hat w)}\int_{-\infty}^{\infty} xf_{\delta}(x)dx = 1 - \mu_{\delta}\frac{f'_{{\textbf{W}}}(\hat w)}{f_{{\textbf{W}}}(\hat w)},
\end{align}
where $\mu_{\delta}$ is the mean of the quantization error $\delta$. Then
\begin{align}
    &\ln \left (\frac{f_{\hat{\textbf{W}}}(\hat w)}{f_{{\textbf{W}}}(\hat w)} \right ) \approx 
    \ln \left ( 1 - \mu_{\delta}\frac{f'_{{\textbf{W}}}(\hat w)}{f_{{\textbf{W}}}(\hat w)} \right )\\\nonumber 
    &\stackrel{|\mu_{\delta}| \text{ is small}}{\approx} - \mu_{\delta}\frac{f'_{{\textbf{W}}}(\hat w)}{f_{{\textbf{W}}}(\hat w)} 
    \label{eq:dev1_kl}
\end{align}
By plugging the equation \eqref{eq:dev1_kl} in KL divergence, we have
\begin{align}
    &\mathbb{D}_{\text{KL}}( f_{{\textbf{W}}}\| f_{\hat{\textbf{W}}})= -\sum_{i} f_{{\textbf{W}}}(\hat w_i) \ln \left (\frac{f_{\hat{\textbf{W}}}(\hat w_i)}{f_{{\textbf{W}}}(\hat w_i)} \right ) \\\nonumber
    &\approx \mu_{\delta}  \sum_{i} f'_{{\textbf{W}}}(\hat w_i).
\end{align}
Then, it follows from Taylor's expansion around the original weight $w_i$, i.e.  $f'_{{\textbf{W}}}(\hat w_i) \approx f'_{{\textbf{W}}}( w_i) + f''_{{\textbf{W}}}( w_i) (\hat w_i - w_i) $ that 
\begin{align}
    \mathbb{D}_{\text{KL}}( f_{{\textbf{W}}}\| f_{\hat{\textbf{W}}}) \approx \mu_{\delta} \sum_{i} f'_{{\textbf{W}}}(w_i) + \mu_{\delta} \sum_{i} f''_{{\textbf{W}}}( w_i) (\hat w_i - w_i),
    \label{eq:dev2_kl}
\end{align}
which proves \textbf{Claim 1}.

To prove \textbf{Claim 2}, 
since $w_i$ and $f''(w_i)$ are bounded, i.e. $|w_i| \leq A$ and $|f''(w_i)| \leq B$ in which $A$ and $B$ are constants, using triangular inequality
\begin{align}
    \nonumber&|\mu_{\delta} \sum_{i} f''_{{\textbf{W}}}( w_i) (\hat w_i - w_i)|= \\
    &|\mu_{\delta}| \sum_{i} |w_i| |f''_{{\textbf{W}}}( w_i)| \left | \left  (\frac {\hat w_i} {w_i} - 1 \right )\right | \leq C \left (\sum_{i} \left |\frac{\hat w_i}{w_i} \right | + 1 \right ),
    \label{eq:dev3_kl}
\end{align}
in which $C = |\mu_{\delta} |AB$. This completes the proof.

Since in PTQ, original weights, and their distribution are known, the first term in equation \eqref{eq:dev2_kl} is constant. Therefore, minimizing $\mathbb{D}_{\text{KL}}( f_{{\textbf{W}}}\| f_{\hat{\textbf{W}}})$ is almost like minimizing $\mu_{\delta} \sum_{i} f''_{{\textbf{W}}}( w_i) (\hat w_i - w_i)$. Thus, following inequality \eqref{eq:dev3_kl}, we may replace $\mathbb{D}_{\text{KL}}( f_{{\textbf{W}}}\| f_{\hat{\textbf{W}}})$ with  $\sum_{i} |\frac{\hat w_i}{w_i}|$ in minimization problem \eqref{eq:obj_quant_KL} which shows Adaptive LASSO is a proxy solution to minimization problem \eqref{eq:obj_quant_KL}.

\section{Experimental Results}\label{sec:experiment}

This section provides experimental results supporting our proposed methodology for pre-calibration quantized LLMs. Note that in our results,  we use a row-wise group quantization technique in conjunction with the soft-thresholding method as explained in Algorithm \ref{alg:al_alg}.

\textbf{Average bits:} The average bit presented in the results of our proposed pre-calibration is calculated based on three factors, (i) the number of non-salaient weights and their bit-width, (ii) the number of salient weights and their bitwidth and (iii) location index of the salient weights.
Since pre-calibration classifies the salient weights in an unstructured manner, tracking the location index is essential to deal with quantized salient and non-salient weights separately.
While maintaining a mask is straightforward, it would add an extra bit per weight, which is inefficient in terms of memory consumption. 
To tackle this issue, we chose to retain the location index of salient weights within each group when using group quantization.
Retaining the index of salient weights leads to a lower average bit since in our method the salient weight ratio $\alpha$, is at most 10\%. This approach results in fewer bits compared to using a mask, i.e. it requires $\log_2 g$ bits only for each salient weight where $g$ is the group size. Moreover, we store scales and zero-points in 16-bit floating-point format. In summary, the average number of bits per weight is computed as
\small
{
\begin{align}
    &b_{\text{avg}}=\\\nonumber
    &\left (b_{\text{C}}+\dfrac{2\times 16}{g} \right)\times (1-\alpha )+ \left (b_{\text{O}}+\log_2 g +\dfrac{2\times 16}{g} \right)\times \alpha
\end{align}
}
where $g$ is the group size, $\alpha$ is the percentage of outlier weights, and $b_{\text{O}}$ and $b_{\text{C}}$ are the bit-widths of outlier and non-outlier weights respectively.

\textbf{Clipping Non-outlier Weights:}
We also used clipping to further reduce the $b_{\text{avg}}$ while maintaining the accuracy in our 3-bit results. The clipping is done because in 3-bit quantization, maintaining quantization accuracy requires a higher ratio of salient weights. On the other hand, increasing the ratio would increase $b_{\text{avg}}$ due to index tracking of outliers. We observed that applying a clipping range of 90-95\% to non-salient weights yields similar accuracy compared to increasing the salient ratio. This confirms that our proposed pre-calibration method can also be combined with other known quantization techniques to achieve better results.


\begin{table}[!b]
    \centering
  
  \scalebox{0.6}{\begin{tabular}{ccccc}
    \toprule
    Model     & Method  & Avg Bits &  Quantization Time (s) $\downarrow$ \\
    \midrule
                 & AWQ (g128) & 4.25  & 838 \\
    LLaMA-7B     & SpQR & 4.63 &  10901  \\
                 & Pre-calibration (g128, $\alpha$=8\%)  & 4.81 &  \textbf{57} \\
    \midrule
                 & AWQ (g128) & 4.25  & 1608   \\
    LLaMA-13B    & SpQR & 4.63 & 20502   \\
                 & Pre-calibration (g128, $\alpha$=6\%)  & 4.67 & \textbf{116}  \\
    \midrule
                 & AWQ (g128) &  4.25 &  3740   \\
     LLaMA-30B   & SpQR &  4.63 &  24069   \\
                 & Pre-calibration (g128, $\alpha$=5\%) & 4.60 & \textbf{470}  \\
    \bottomrule
  \end{tabular}}
    \caption{Quantization time comparison}\label{tab:time}
  
\end{table}

\textbf{Perplexity:} We evaluated perplexity of quantized LLaMA models on WikiText2 \cite{merity2016pointer} and C4 \cite{raffel2020exploring} datasets when sequence length is 2048. Table \ref{tab:perp} shows the results comparing perplexity scores for FP16, Round to Nearest (RTN), GPTQ \cite{optq}, AWQ\cite{awq}, SpQR,\cite{dettmers2024spqr} and our proposed pre-calibration method.
Pre-calibration outperforms AWQ and RTN consistently in terms of perplexity scores. Furthermore, pre-calibration exhibits perplexity scores that closely follow those of SpQR and OmniQuant, particularly for larger models. These results highlight the pre-calibration ability to achieve competitive accuracy while offering a significant advantage in terms of quantization time efficiency and robustness. These results can be used as an initial point to further optimize the model using calibration. Refer to Appendix \ref{app:extra-results} for more perplexity results on Falcon \cite{falcon40b} and OPT
\cite{zhang2022opt} models.

\textbf{Quantization Time:} Benefiting from our simple soft-thresholding technique, our proposed pre-calibration method significantly reduces the quantization time compared to existing methods. The proposed method achieves at least 10$\times$ faster quantization speed than AWQ \cite{awq} and surpasses SpQR \cite{dettmers2024spqr} quantization time by a factor of 100$\times$ as shown in Table~\ref{tab:time}. 
%


\textbf{Zero-Shot Task Evaluation:} We also evaluated the accuracy of LLaMA 1 \cite{touvron2023llama} and  LLaMA 2 \cite{touvron2023llama2} models on 5 zero-shot common-sense reasoning tasks including ARC(easy and challenge) \cite{clark2018think}, HellaSwag \cite{zellers2019hellaswag}, WinoGrande \cite{sakaguchi2021winogrande} and PIQA \cite{bisk2020piqa} using LM Evaluation Harness \cite{gao2021framework}. As shown in Table \ref{tab:lmeval}, our proposed pre-calibration outperforms SpQR \cite{dettmers2024spqr} in both 4-bit and 3-bit quantization, showing correctly classifying salient weight in pre-calibration step can improve the quality of PTQ to a great extent.

\begin{table*}[!t]
  \centering
  \caption{Comparison of the pre-calibration perplexity results of 4-bit \& 3-bit on WikiText2. }\label{tab:perp}
  \scalebox{0.6}{
  \begin{tabular}{ccccc||ccc}
  \toprule
  &&\multicolumn{3}{c}{\textbf{4-bit}}&\multicolumn{3}{c}{\textbf{3-bit}} \\
  \cmidrule(r){3-5} \cmidrule(r){6-8}\\
    Model     & Method & Quantization setting & Avg Bits & Wiki2 $\downarrow$    & Quantization setting & Avg Bits & Wiki2 $\downarrow$     \\
    \hline
    \multirow{7}{*}{\rotatebox{90}{\textbf{\small{LLaMA-7B}}}}
                & FP16 &- &  16.00 & 5.67  &-  &  16.00 & 5.67 \\
                & RTN  & 4bit-g128  &  4.25 & 5.96 & 3bit-g128 & 3.25 & 7.01  \\
                & OPTQ & 4bit-g128 & 4.25&
                5.83 &3bit-g128 & 3.25 & 6.58  \\
               & AWQ  &4bit-g128  & 4.25 & 5.78 &3bit-g128 & 3.25 & 6.35    \\
               & OmniQuant  &4bit-g128  & 4.25 &5.77 &3bit-g128 & 3.25 & 6.15   \\
                & SpQR & Refer to Appendix \ref{app:hp_config}  & 4.63 & \textbf{5.73} &Refer to Appendix \ref{app:hp_config}   & 3.98 & \textbf{5.87}  \\
                & Pre-calibration &(4bit-g128, $\alpha=8\%$)   & 4.81 & 5.78 &(3bit-g128, $\alpha=9\%$, $b_\text{O}$=4)  & 3.97 & 6.07  \\
    \hline
    \multirow{7}{*}{\rotatebox{90}{\textbf{\small{LLaMA-13B}}}}
               & FP16 &-  &  16.00 & 5.09  & - &   16.00 & 5.09   \\
                & RTN  &4bit-g128 &  4.25 & 5.25 &3bit-g128   & 3.25 & 5.88     \\
                & OPTQ & 4bit-g128 &4.25&
                5.20 &3bit-g128 & 3.25 & 5.70\\
              & AWQ &4bit-g128  &4.25 & 5.18 &3bit-g128  & 3.25 & 5.52  \\   
              & OmniQuant  &4bit-g128  & 4.25 &5.17 &3bit-g128 & 3.25 & 5.44    \\
                & SpQR &Refer to Appendix \ref{app:hp_config}   & 4.63 & \textbf{5.13}  &Refer to Appendix \ref{app:hp_config}   & 3.97 & \textbf{5.22}    \\    
                & Pre-calibration & (4bit-g128, $\alpha=6\%$)  & 4.67 & \textbf{5.15}  &(3bit-g128, $\alpha=9\%$, $b_\text{O}$=4)  & 3.97 & 5.32  \\
    \hline
    \multirow{7}{*}{\rotatebox{90}{\textbf{\small{LLaMA-30B}}}}
               & FP16 &-  &  16.00 & 4.10  &-   &  16.00 & 4.10    \\
                & RTN &4bit-g128  & 4.25 & 4.23 &3bit-g128   & 3.25 &4.88    \\
                & OPTQ & 4bit-g128 &4.25&
                4.22 &3bit-g128 & 3.25 & 4.74  \\
                & AWQ  &4bit-g128  & 4.25 & 4.21  &3bit-g128  & 3.25 & 4.61 \\
                & OmniQuant  &4bit-g128  & 4.25 &4.19  &3bit-g128 & 3.25 & 4.56   \\
                & SpQR &Refer to Appendix \ref{app:hp_config}  &  4.63 & \textbf{4.14}  &Refer to Appendix \ref{app:hp_config}    & 3.90 & \textbf{4.25}  \\
                & Pre-calibration &(4bit-g128, $\alpha=5\%$)   & 4.60 & \textbf{4.16}  &(3bit-g128, $\alpha=8\%$, $b_\text{O}$=4)  & 3.89 & \textbf{4.31}   \\
    \hline
    \multirow{7}{*}{\rotatebox{90}{\textbf{\small{LLaMA2-7B }}}}
                 & FP16 &-  & 16.00 & 5.47&-  & 16.00 & 5.47   \\
                 & RTN  &4bit-g128  & 4.25 & 5.72  &3bit-g128   & 3.25 &6.66    \\
                 & OPTQ & 4bit-g128 &4.25&
                5.61  &3bit-g128 & 3.25 & 6.38\\
                 & AWQ &4bit-g128 & 4.25 & 5.60  &3bit-g128  & 3.25 & 6.24  \\
                 & OmniQuant  &4bit-g128  & 4.25 &5.58&3bit-g128 & 3.25 &6.03  \\
                 & SpQR &Refer to Appendix \ref{app:hp_config}  &  4.63 & \textbf{5.52}  &Refer to Appendix \ref{app:hp_config}   & 3.98 & \textbf{5.66}  \\
                 & Pre-calibration  &(4bit-g128, $\alpha=8\%$) & 4.81 & 5.60  &(3bit-g128, $\alpha=9\%$, $b_\text{O}$=4)  & 3.97 & 5.83  \\
    \hline
    \multirow{7}{*}{\rotatebox{90}{\textbf{\small{LLaMA2-13B}}}}
                  & FP16 &-   & 16.00 & 4.88  &-  & 16.00 & 4.88  \\
                  & RTN &4bit-g128   & 4.25 & 4.98  &3bit-g128  & 3.25 & 5.52   \\
                  & OPTQ & 4bit-g128 &4.25&
                4.99  &3bit-g128 & 3.25 & 5.42\\
                  & AWQ  &4bit-g128  & 4.25 &4.97 &3bit-g128 & 3.25 & 5.32    \\
                  & OmniQuant  &4bit-g128  & 4.25 &4.95  &3bit-g128 & 3.25 & 5.28  \\
                  & SpQR  & Refer to Appendix \ref{app:hp_config} & 4.63 & \textbf{4.92}   &Refer to Appendix \ref{app:hp_config} & 3.96 & \textbf{5.01}   \\
                  & Pre-calibration  &(4bit-g128, $\alpha=6\%$) & 4.67 & \textbf{4.93}  &(3bit-g128, $\alpha=9\%$, $b_\text{O}$=4)  & 3.97 & \textbf{5.05}     \\
    \bottomrule
  \end{tabular}}
    
\end{table*}
\begin{table*}[!t]
\caption{Comparison of the pre-calibration results on zero-shot tasks using LM Evaluation Harness \cite{gao2021framework}.}\label{tab:lmeval}
  \centering
  \scalebox{0.70}{
  \begin{tabular}{cccccccccc}
    \toprule
    Model     && Method & Avg Bit & ARC-c & ARC-e & HellaSwag   & Winogrande & PIQA  & Avg\\
    \midrule
    \multirow{6}{*}{\rotatebox{90}{\textbf{LLaMA-7B}}}
                && FP16 &16 & 41.89 & 75.25 & 56.95 & 69.93 & 78.67 & 64.54 \\
                && RTN (g128) & 4.25 & \textbf{42.92} & \textcolor{gray}{74.54} & \textcolor{gray}{56.29} & 70.01 & \textcolor{gray}{78.18} & \textcolor{gray}{64.39}  \\
                && OPTQ (g128) & 4.25 & \textcolor{gray}{40.78} & \textcolor{gray}{74.62} & \textcolor{gray}{56.59} & \textcolor{gray}{69.22} & \textcolor{gray}{78.51} & \textcolor{gray}{63.94}  \\
                && AWQ (g128) & 4.25 & \textcolor{gray}{41.13} & \textcolor{gray}{75.00} & \textcolor{gray}{56.44} & \textcolor{gray}{69.14} & \textcolor{gray}{77.86} & \textcolor{gray}{63.91}  \\
                && SpQR$^*$ & 4.63& \textcolor{gray}{41.72} & 75.21 & 56.65 & \textcolor{gray}{69.61} & \textbf{79.05} & 64.45 \\
                && Pre-calibration (g128, $\alpha=8\%$)& 4.81 & 42.15 & \textbf{75.34} & \textbf{56.72} & \textbf{70.17} & 78.56 & \textbf{64.59} \\
    \midrule
    \multirow{6}{*}{\rotatebox{90}{\textbf{LLaMA-13B }}}
               && FP16 &16 & 46.42 & 77.36 & 59.88 & 72.69 & 79.16 &  67.19 \\
                && RTN (g128)  & 4.25 & 45.82 & \textcolor{gray}{76.77} & \textcolor{gray}{59.37} & \textcolor{gray}{72.45} & \textbf{79.71} & \textcolor{gray}{66.82}  \\
                && OPTQ (g128)  & 4.25 & \textbf{45.99} & \textbf{77.06} & \textcolor{gray}{59.22} & \textbf{73.32} & \textcolor{gray}{78.94} & 66.91  \\
                && AWQ (g128)  & 4.25 & \textbf{45.99} & \textcolor{gray}{76.89} & 59.42 & \textcolor{gray}{72.53} & \textcolor{gray}{78.78} & \textcolor{gray}{66.72}  \\
                 && SpQR$^*$ & 4.63  & \textcolor{gray}{45.73} & \textcolor{gray}{76.85} & \textbf{59.70} & 73.09 & 79.22   & \textbf{66.92} \\
                && Pre-calibration (g128, $\alpha=6\%$) & 4.67 & \textbf{45.99} & \textcolor{gray}{76.85} & 59.41 & \textcolor{gray}{73.01} & \textcolor{gray}{78.94} & \textcolor{gray}{66.84} \\
    \midrule
    \multirow{6}{*}{\rotatebox{90}{\textbf{LLaMA-30B }}}
                && FP16 &16  & 52.90 & 80.43 & 63.37 & 75.85 & 81.12 & 70.73  \\
                && RTN (g128)& 4.25& 52.05 & \textbf{80.77} & \textcolor{gray}{62.89} & \textcolor{gray}{74.19} & \textcolor{gray}{80.58}  & \textcolor{gray}{70.10}  \\
                && OPTQ (g128)& 4.25& \textcolor{gray}{51.37} & \textcolor{gray}{80.47} & 63.12 & 75.30 & \textbf{80.79}  & 70.21  \\
                && AWQ (g128)& 4.25& \textbf{53.41} & 80.72 & \textbf{63.16} & \textbf{75.45} & \textcolor{gray}{80.69}  & \textbf{70.69}  \\
                && SpQR$^*$ & 4.63 & \textcolor{gray}{51.45} & \textcolor{gray}{80.47} & \textcolor{gray}{63.08} & \textcolor{gray}{74.74} & 80.74 & \textcolor{gray}{70.10}    \\
                && Pre-calibration (g128, $\alpha=5\%$)& 4.60 & \textcolor{gray}{51.88} & \textbf{80.77} & \textcolor{gray}{63.07} & \textcolor{gray}{74.19} & 80.74  & \textcolor{gray}{70.13}    \\
    \midrule
    \multirow{6}{*}{\rotatebox{90}{\textbf{LLaMA-2-7B}}}
                 && FP16 &16  & 43.43 & 76.35 & 57.16 & 69.14 & 78.07 & 64.83   \\
                && RTN (g128)& 4.25& \textcolor{gray}{43.09} & 76.18 & \textcolor{gray}{56.90} & \textcolor{gray}{68.67} & \textcolor{gray}{77.48} & \textcolor{gray}{64.46}  \\
                && OPTQ (g128) & 4.25 & \textcolor{gray}{41.89} & \textcolor{gray}{74.96} & \textcolor{gray}{56.33} & 69.30 & \textbf{77.97} & \textcolor{gray}{64.09}  \\
                && AWQ (g128) & 4.25 & \textcolor{gray}{42.58} & \textcolor{gray}{75.67} & \textcolor{gray}{56.39} & \textcolor{gray}{68.35} & 77.53 & \textcolor{gray}{64.10}  \\
                && SpQR$^*$ & 4.63  & \textbf{44.28} & \textcolor{gray}{76.14} & 56.95 & \textcolor{gray}{68.51} & \textcolor{gray}{77.42} & 64.66  \\
                && Pre-calibration (g128, $\alpha=8\%$)& 4.81 & 43.17 & \textbf{76.39} & \textbf{57.12} & \textbf{69.77} & \textbf{77.97} &  \textbf{64.88} \\
    \midrule
    \multirow{6}{*}{\rotatebox{90}{\textbf{LLaMA-2-13B}}}
                && FP16 &16  & 48.46 & 79.42 & 60.05 & 72.38 & 79.11 & 67.88  \\
                && RTN  (g128)& 4.25& \textcolor{gray}{48.12} & \textcolor{gray}{78.83} & \textcolor{gray}{59.74} & \textcolor{gray}{72.69} & \textcolor{gray}{78.67}  & \textcolor{gray}{67.61}  \\
                && OPTQ  (g128)& 4.25& \textcolor{gray}{47.95} & \textcolor{gray}{78.79} & \textcolor{gray}{59.81} & 72.85 & \textcolor{gray}{78.56}  & \textcolor{gray}{67.60}  \\
                && AWQ  (g128)& 4.25& \textcolor{gray}{46.59} & \textcolor{gray}{79.46} & \textcolor{gray}{59.85} & \textbf{73.32} & \textbf{79.05}  & \textcolor{gray}{67.65}  \\
                && SpQR$^*$ & 4.63 & \textbf{48.46} & \textbf{79.76} & \textbf{59.97} & \textcolor{gray}{71.90} & \textcolor{gray}{78.84}  & 67.79  \\
                && Pre-calibration (g128, $\alpha=6\%$)& 4.67 & 48.38 & 79.63 & 59.89 & \textcolor{gray}{72.53} & 78.94 & \textbf{67.87}   \\
    \bottomrule
  \end{tabular}
  }

\small{$^*$ Refer to Appendix \ref{app:hp_config} for quantization settings.}
\end{table*}

\section{Diverging View on Improving Post-training Quantization}

Traditional post-training quantization (PTQ) methods typically follow a two-step process: quantization followed by calibration. While this approach has been effective, the calibration step often poses significant challenges, as it involves adjusting the quantized weights to recover as much of the original model's accuracy as possible. The difficulty of this task is compounded by the fact that calibration is inherently a complex optimization problem, and based on optimization theory, having a better initial point can greatly influence the outcome.

In this paper, we propose a shift in perspective by introducing a pre-calibration step prior to the traditional calibration process. We have demonstrated that pre-calibration can provide a significantly improved starting point for calibration, enhancing the overall effectiveness of the PTQ process. In fact, in some cases, this pre-calibration starting point outperforms even the final results of previously introduced calibration methods.

Another critical aspect of our approach is the \textit{classification} of weights during quantization. While we utilized KL divergence in conjunction with Adaptive LASSO, where the Adaptive LASSO serves as a proxy solution for minimizing KL divergence, this framework is flexible. The choice of divergence measure or regularization penalty is not fixed; one could employ other $f$-divergence measures or adapt different penalties based on the specific needs of the PTQ method. Our proposal is not prescriptive in this regard but rather encourages the exploration of the best tools for achieving optimal \textit{weight classification} before the PTQ procedure.

\section{Conclusion}

We presented a weight-adaptive pre-calibration approach for PTQ methods. Traditional PTQ techniques typically rely on a two-step process of quantization followed by calibration. However, the calibration step often proves challenging, as it requires careful adjustments to quantized weights to regain the model's original accuracy. 
We have demonstrated that pre-calibration can provide a significantly improved starting point for calibration, enhancing the overall effectiveness of the PTQ process and, in some cases, even surpassing the final performance of previously introduced calibration methods. This highlights the importance of starting with a well-prepared initial point, which can significantly impact the success of the quantization process.
Our work rethinks the traditional PTQ pipeline, advocating for the integration of a pre-calibration step that enhances the starting conditions for calibration. This shift not only improves the robustness and effectiveness of the quantization process but also opens the door to further innovations in model efficiency and deployment strategies. As the demand for efficient deployment of Large Language Models (LLMs) continues to grow, our approach provides a new perspective on optimizing PTQ for diverse applications.

\bibliographystyle{apalike}
{\small
\bibliography{example}}

\section*{\uppercase{Appendix}}

\section{Experimental Settings }\label{app:settings}

\subsection{Seed Sensitivity}\label{app:seed}
Since our proposed method, only uses deterministic pre-trained weights of the model and performs a soft-thresholding to identify $\alpha$ percent of outlier weights, it does not exhibit any stochastic behavior during the quantization. Furthermore, we do not use any data for calibration and thus, our algorithm is robust toward randomness in data selection. We believe this is the main advantage of our proposed algorithm.

\subsection{Calibration Datasets and Parameters }
We follow the pipelines used in SpQR\footnote{See \url{https://github.com/Vahe1994/SpQR}} and AWQ\footnote{See \url{https://github.com/mit-han-lab/llm-awq}} official implementation to generate calibration datasets. Random selection of 128 samples of length 2048 form RedPajama \cite{together2023redpajama}, C4 and RefinedWeb \cite{refinedweb} is used for quantization of LLaMA 1, LLaMa 2, OPT \cite{zhang2022opt} and Falcon \cite{falcon40b} using SpQR. For AWQ experiments 128 samples from a small subset of Pile \cite{gao2020pile} dataset is used following the AWQ's implementation.

\subsection{Hyper-Parameters and Configs} \label{app:hp_config}

\textbf{RTN:} We implemented RTN quantization method based on the implementation of AWQ which supports weight reshaping for group quantization.

\noindent\textbf{AWQ:} We used AWQ's official implementation for quantizing LLaMA and OPT models.

\noindent\textbf{SpQR:} We use SpQR's official implementation for quantizing LLaMA, Code-Llama and OPT models. Table \ref{tab:spqr config} shows the hyper-parameters used for SpQR quantization.

\begin{table}[!h]
    \centering
  \caption{Quantization configuration of SpQR}
  \label{tab:spqr config}
  \scalebox{0.6}{\begin{tabular}{cccccc}
    \toprule
    Model     & Calibration & Group &  Weight & Scales \& Zeros & Outlier  \\
    & Set &Size & Bits & Bits & Threshold\\
    \midrule
    LLaMA     & RedPajama & 16 &  4 & 3& 0.2  \\
           & RedPajama & 16 &  3 & 3& 0.25-0.28  \\
    \midrule
    Code-Llama     & RedPajama & 16 &  4& 3 & 0.2  \\
    \midrule
    OPT     & C4 & 16 &  4 & 3& 0.2  \\
           \midrule
    Falcon     & RefinedWeb & 16 &  4 & 3& 0.2  \\
    \bottomrule
  \end{tabular}}
  
\end{table}

\subsection{Hardware Settings}\label{app:hw}
We perform quantization on single NVIDIA V100-32G GPU. For evaluation using LM Evaluation Harness we use 8$\times$V100-32G GPUs for 30B models.

\section{Extra Experimental Results}\label{app:extra-results}
Table \ref{tab:perp opt} shows the perplexity results on OPT \cite{zhang2022opt} and Falcon\cite{falcon40b} models.

\begin{table}[!h]
  \caption{Perplexity of 4-bit OPT and Falcon models on WikiText2 and C4.}
  \label{tab:perp opt}
  \centering
  \scalebox{0.6}{
  \begin{tabular}{cccccc}
  \toprule
    Model     & Method & Quantization setting & Avg Bits & Wiki2 $\downarrow$ & C4$\downarrow$   \\
    \hline
    \multirow{5}{*}{\rotatebox{0}{\textbf{\small{OPT-6.7B}}}}
                & FP16 &- &  16.00 & 10.86 & 11.74 \\
                & RTN  & 4bit-g128  &  4.25 & 11.15 & 12.31  \\
       & AWQ  &4bit-g128  & 4.25 & 10.95 & 11.86    \\
                & SpQR & Refer to Appendix \ref{app:hp_config}  & 4.63 & 10.91 & 11.78  \\
                & Pre-calibration &(4bit-g128, $\alpha=6\%$)   & 4.67 & \textbf{10.86} & 11.99 \\
    \hline
    \multirow{5}{*}{\rotatebox{0}{\textbf{\small{OPT-13B}}}}
               & FP16 &-  &  16.00 & 10.13 & 11.20\\
                & RTN  &4bit-g128 &  4.25 & 10.30 & 11.51    \\
              & AWQ &4bit-g128  &4.25 & 10.29 & 11.28   \\
                & SpQR &Refer to Appendix \ref{app:hp_config}   & 4.27 & \textbf{10.22} & \textbf{11.27}  \\    
                & Pre-calibration & (4bit-g128, $\alpha=6\%$)  & 4.67 & \textbf{10.20} & \textbf{11.31}  \\
    \hline
    \multirow{5}{*}{\rotatebox{0}{\textbf{\small{OPT-30B}}}}
               & FP16 &-  &  16.00 & 9.55 & 10.69    \\
                & RTN &4bit-g128  & 4.25 & 9.94 & 10.94   \\
                & AWQ  &4bit-g128  & 4.25 & 9.61 & 10.74  \\
                & SpQR &Refer to Appendix \ref{app:hp_config}  &  4.63 & 9.55 & 10.71  \\
                & Pre-calibration &(4bit-g128, $\alpha=5\%$)   & 4.60 & 9.64 & 10.79    \\
    \hline
    \multirow{5}{*}{\rotatebox{0}{\textbf{\small{  Falcon-7B}}}}
                 & FP16 &-  & 16.00 & 6.59 & 9.50  \\
                 & RTN  &4bit-g128  & 4.25 & 6.79 & 9.79   \\
                 & SpQR &Refer to Appendix \ref{app:hp_config}  &  4.44 & \textbf{6.64} & \textbf{9.58}\\
                 & Pre-calibration  &(4bit-g128, $\alpha=4\%$) & 4.53 & \textbf{6.69} & \textbf{9.63}  \\
    \hline
    \multirow{5}{*}{\rotatebox{0}{\textbf{\small{ Falcon-40B}}}}
                  & FP16 &-   & 16.00 & 5.23& 7.76   \\
                  & RTN &4bit-g128   & 4.25 & 5.31 & 7.88   \\
                  & SpQR  &Refer to Appendix \ref{app:hp_config} & 4.46 & \textbf{5.26} & \textbf{7.79}  \\
                  & Pre-calibration  &(4bit-g128, $\alpha=5\%$) & 4.60 &\textbf{5.27} & \textbf{7.81}   \\
    \bottomrule
  \end{tabular}}
\end{table}

\end{document}